\RequirePackage{fix-cm}
\documentclass[twocolumn]{svjour3}
\def\makeheadbox{} 

\usepackage[utf8]{inputenc}
\usepackage[american]{babel} 
\usepackage{graphicx}
\usepackage{mathptmx}	
\usepackage{latexsym}
\usepackage{url}
\usepackage[pdftex,unicode]{hyperref}
\usepackage{amsmath,amssymb,amsfonts}

\newcommand\wrt  {w.r.t.}

\newcommand\eg   {e.g.}
\newcommand\PROLOG    {\index{Prolog@{\sc Prolog}}{\sc Prolog}}
\newcommand\ie   {i.e.}

\newcommand\namefont        {\rm}
\newcommand\david           {{\namefont David}}
\newcommand\poole           {{\namefont Poole}}
\newcommand\poolename       {{\namefont\david\ \poole}}

\newcommand\nthpositioner[2]
{#1\raisebox{0.52ex}{\scriptsize\hspace{0.07em}#2}}

\newcommand\mthpositioner[2]
{\math{#1}\raisebox{0.52ex}{\scriptsize\hspace{0.07em}#2}}
\newcommand\modulointocountzero[2]
{\count1=#1
\count2=#2
\count0=\count1
\divide  \count0 by \count2
\multiply\count0 by-\count2
\advance \count0 by \count1}
\newcommand\absolutevalueintocountzero[1]
{\count0=#1
\ifnum\count0<0\multiply\count0 by -1\fi}
\newcommand\nthstring[1]
{\def\myargone{#1}\ifcat a\myargone th\else\nthstringnochar{#1}\fi}
\newcommand\nthstringnochar[1]
{\absolutevalueintocountzero{#1}%
\modulointocountzero{\count0}{100} 
\ifnum\count0>9\ifnum\count0<20 th\else\stupidnthstring\fi
                                  \else\stupidnthstring\fi}
\newcommand\stupidnthstring
{\modulointocountzero{\count0}{10}
\ifnum\count0=1 \hskip-0.2em st\else
\ifnum\count0=2 nd\else
\ifnum\count0=3 rd\else 
                th\fi\fi\fi}

\newcommand\writeascents
[1]{\count4=#1
\ifnum\count4<0 
-\multiply\count4 by -1\fi
\modulointocountzero{\count4}{10}
\divide\count4 by 10
\count3=\the\count0
\modulointocountzero{\count4}{10}
\divide\count4 by 10
\the\count4
.\the\count0
\the\count3
}

\newcommand\frenchnthstring[1]
{\def\myargone{#1}\ifcat a\myargone th\else\frenchnthstringnochar{#1}\fi}
\newcommand\frenchnthstringnochar[1]
{\absolutevalueintocountzero{#1}%
\modulointocountzero{\count0}{100} 
\ifnum\count0>9\ifnum\count0<20 th\else\frenchstupidnthstring\fi
                                  \else\frenchstupidnthstring\fi}
\newcommand\frenchstupidnthstring
{\modulointocountzero{\count0}{10}
\ifnum\count0=1 \hskip-0.2em re\else
\ifnum\count0=2 me\else
\ifnum\count0=3 rd\else 
                th\fi\fi\fi}

\journalname{Künstliche Intelligenz}

\usepackage{picinpar}
\newcommand{\cv}[3]{\bigskip\par\noindent
\begin{window}[0,l,{\includegraphics[width=20mm]{#1}},{}]
  \noindent
  \textbf{#2} #3
\end{window}\par
}

\begin{document}

\title{The RatioLog Project: Rational Extensions of Logical Reasoning}
\author{Ulrich Furbach \and Claudia Schon \and Frieder Stolzenburg \and Karl-Heinz Weis \and Claus-Peter Wirth}
\institute{U. Furbach, C. Schon, K.-H. Weis \at
	    Artificial Intelligence Research Group,
	    Universität Koblenz-Landau,
	    Universitätsstr. 1,
	    56070~Koblenz, Germany\\
            \email{\{uli, schon\}@uni-koblenz.de, khweis@weis-consulting.de}
	\and
           F. Stolzenburg, C.-P. Wirth \at
              Automation and Computer Sciences Department,
              Harz University of Applied Sciences,
              Friedrichstr. 57--59,
              38855~Wernigerode, Germany\\
              \email{fstolzenburg@hs-harz.de, wirth@logic.at}
}
\date{}

\maketitle

\begin{abstract}
Higher-level cognition includes logical reasoning and the ability of question
answering with common sense. The RatioLog project addresses the problem of
rational reasoning in deep question answering by methods from automated
deduction and cognitive computing. In a first phase, we combine techniques from
information retrieval and machine learning to find appropriate answer candidates
from the huge amount of text in the German version of the free encyclopedia
``Wikipedia''. In a second phase, an automated theorem prover tries to verify
the answer candidates on the basis of their logical representations. In a third
phase \mbox{---~because} the knowledge may be incomplete and inconsistent~---,
we consider extensions of logical reasoning to improve the results. In this
context, we work toward the application of techniques from human reasoning: We
employ defeasible reasoning to compare the answers \wrt\ specificity, deontic
logic, normative reasoning, and model construction. Moreover, we use integrated
case-based reasoning and machine learning techniques on the basis of the
semantic structure of the questions and answer candidates to learn giving the
right answers.
\keywords{automated deduction \and
case-based reasoning \and 
common-sense reasoning \and 
defeasible reasoning \and
deontic logic \and
question answering\and 
specificity}
\end{abstract}

\section{Rational Reasoning and Question Answering}\label{introduction}
\begin{figure*}[t]
\centering
\includegraphics[width=\textwidth]{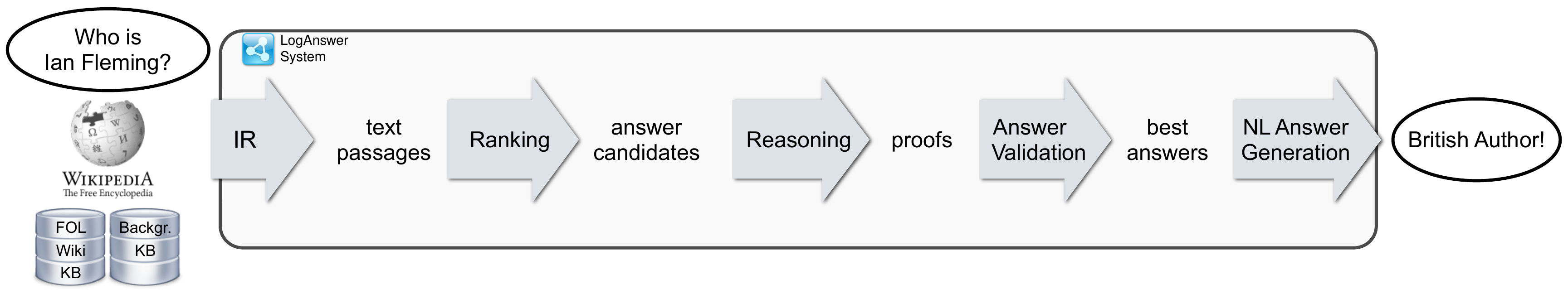}
\caption{The LogAnswer system uses information retrieval (IR), decision tree learning in a ranking phase, reasoning, and natural language answer generation to compute answers.}
\label{fig:architecture}
\end{figure*}

The development of formal logic played a big role in the field of automated
reasoning, which led to the development of the field of artificial intelligence
(AI)\@. Applications of {\em automated deduction}\/ in mathematics have been
investigated from the early years on. Nowadays automated deduction techniques
are successfully applied in hard- and software verification and many other areas
(for an overview see \cite{DBLP:journals/expert/BeckertH14}).

In contrast to formal logical reasoning, however, human reasoning does not
strictly follow the rules of classical logic. Reasons may be incomplete
knowledge, incorrect beliefs, and inconsistent norms. From the very beginning of
AI research, there has been a strong emphasis on incorporating mechanisms for
rationality, such as abductive or defeasible reasoning. From these efforts, as
part of the field of knowledge representation, \emph{common-sense reasoning} has
emerged as a branching discipline with many applications in AI \cite{Mul14}.

Nowadays there is a chance to join automated deduction and common-sense
reasoning within the paradigm of \emph{cognitive computing}, which allows the
implementation of rational reasoning \cite{Watson2013}. The general motivation
for the development of cognitive systems is that computers can solve
well-defined mathematical problems with enormous precision at a reasonably sufficient
speed in practice. It remains difficult, however, to solve problems that
are only vaguely outlined. One important characteristic of cognitive computing
is that many different knowledge formats and many different information
processing methods are used in a combined fashion. Also the amount of knowledge
is huge and, even worse, it is even increasing steadily. For the logical
reasoning, a similar argument holds: Different reasoning mechanisms have to be
employed and combined, such as classical deduction (forward reasoning) on the
one hand, and abduction or other non-monotonic reasoning mechanisms on the
other hand. 

Let us illustrate this with a well-known example from the literature:

\begin{enumerate}\em
  \item Tom is an emu.
  \item Emus are birds.
  \item Birds normally fly.
  \item Emus do not fly.
\end{enumerate}

The question is: Can emus fly or not? Forward reasoning allows us to infer that
emus are birds and hence can normally fly. This is in conflict, however, with
the strict background knowledge that emus do not fly. The conflict can be solved
by assuming certain knowledge as default or defeasible, which only holds
normally. Hence we may conclude here that emus and therefore Tom does not fly.
We will come back to this example later (namely in Section \ref{specificity}~and~\ref{deontic}).

Rational reasoning must be able to deal with incomplete as well as conflicting
(or even inconsistent) knowledge. Moreover, huge knowledge bases with
inconsistent contents must be handled. Therefore, it seems to be a good idea to
combine and thus enhance rational reasoning by information retrieval techniques,
\eg\ techniques from machine learning. This holds especially for the domain of
deep question answering, where communication with patterns of human reasoning is
desirable.

\subsection{Deep Question Answering and the LogAnswer System}\label
{section Deep Question Answering}

Typically, question answering systems, including application programs such as
\emph{Okay Google}$^{\scriptscriptstyle\circledR}$ or
Apple$^{\scriptscriptstyle\circledR}$'s \emph{Siri}, communicate with the user
in natural language. They accept properly formulated questions and return
concise answers. These automatically generated answers are usually not extracted
directly from the web, but, in addition, the system operates on an extensive
(background) knowledge base, which has been derived from textual sources in
advance.

LogAnswer \cite{Furbach:Gloeckner:Helbig:Pelzer:LogicBasedQA:2009,Fu10} is an
open-domain question answering system, accessible via a web interface
(\url{www.loganswer.de}) similar to that of a search engine. The knowledge used
to answer the question is gained from 29.1 million natural-language sentences of
a snapshot of the German Wikipedia. Furthermore, a background knowledge
consisting of 12,000 logical facts and rules is used. The LogAnswer system was
developed in the DFG-funded LogAnswer project, a cooperation between the groups
on Intelligent Information and Communication Systems at the Fern\-Universität
Hagen and the AI research group at the University of Koblenz-Landau. The project
aimed at the development of efficient and robust methods for logic-based
question answering. The user enters a question and LogAnswer presents the five
best answers from a snapshot of the German ``Wikipedia'', highlighted in the
context of the relevant textual sources.

Most question answering systems rely on shallow linguistic methods for answer
derivation, and there is only little effort to include semantics and logical
reasoning. This may make it impossible for the system to find any answers: A
superficial word matching algorithm is bound to fail if the textual sources use
synonyms of the words in the question. Therefore, the LogAnswer system models
some form of background knowledge, and combines cognitive aspects of linguistic
analysis, such as semantic nets in a logical representation, with machine
learning techniques for determining the most appropriate answer candidate.

Contrary to other systems, LogAnswer uses an automated theorem prover to compute
the replies, namely Hyper \cite{cadesd}, an implementation of the hypertableaux
calculus \cite{DBLP:conf/jelia/BaumgartnerFN96}, extended with equality among
others. It has demonstrated its strength in particular for reasoning problems
with a large number of irrelevant axioms, as they are characteristic for the
setting of question answering. The logical reasoning is done on the basis of a
logical representation of the semantics of the entire text contained in the
Wikipedia snapshot. This is computed beforehand with a system developed by
computational linguists \cite{GH07} which employs the MultiNet graph formalism
(Multilayered Extended Semantic Networks) \cite{He06}.

Since methods from natural-language processing are often confronted with flawed
textual data, they strive toward robustness and speed, but often lack the
ability to perform more complex inferences. By contrast, a theorem prover uses a
sound calculus to derive precise proofs of a higher complexity; even minor flaws
or omissions in the data, however, lead to a failure of the entire derivation
process. Thus, additional techniques from machine learning, defeasible and
normative reasoning etc.\ should be applied to improve the quality of the
answers --- as done in the RatioLog project.

For this, the reasoning in classical logic is extended by various forms of
non-monotonic aspects, such as defeasible argumentation. By these extensions,
the open-domain question answering system LogAnswer is turned into a system for
rational question answering, which offers a testbed for the evaluation of
rational reasoning.

\subsection{The LogAnswer System and its Modules}\label{subsection The LogAnswer System and Hyper}

When processing a question, the LogAnswer system performs several different
steps. Figure~\ref{fig:architecture} presents details on these steps. 
\begin{figure}[t]
\includegraphics[width=8.5cm]{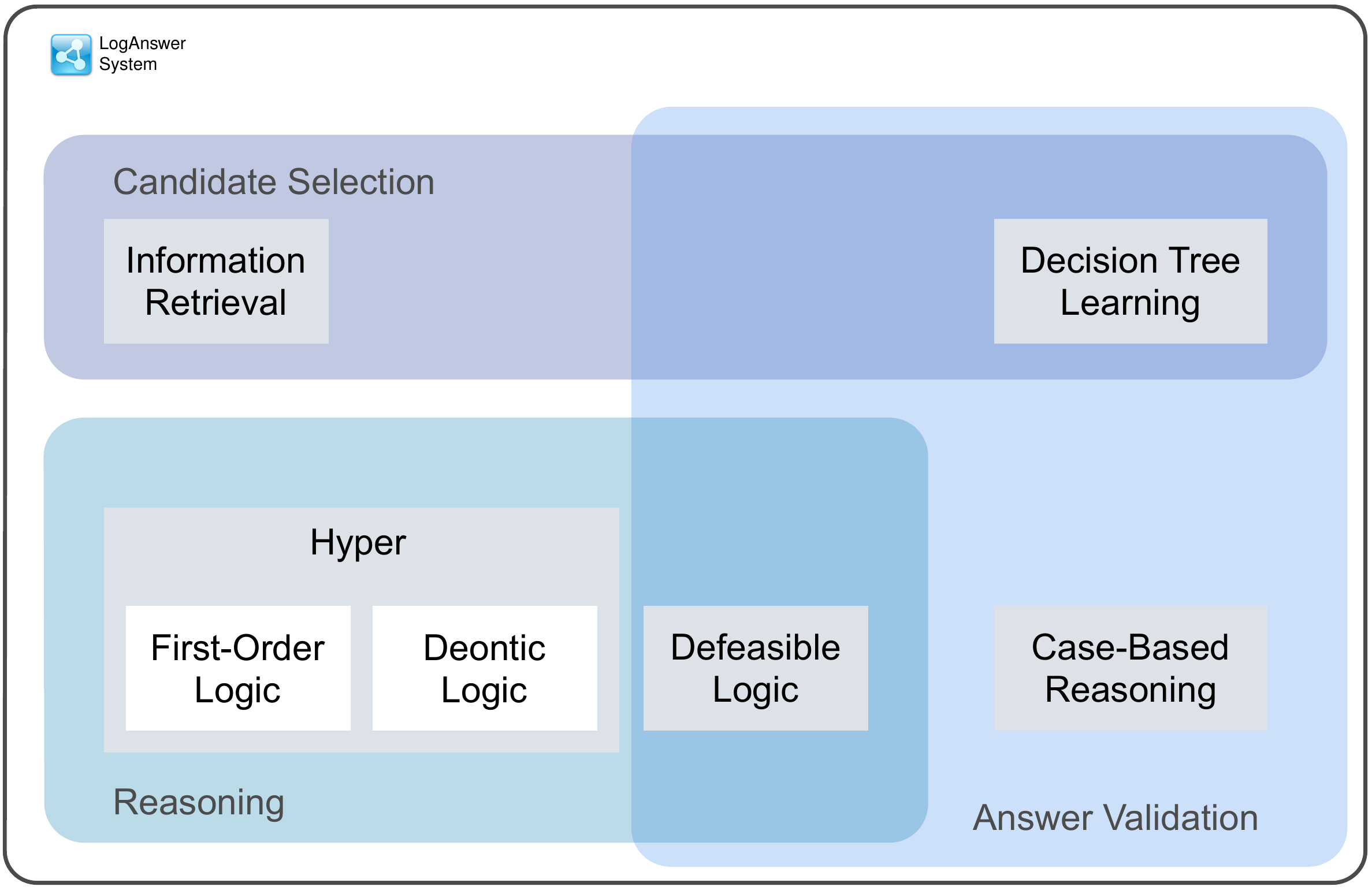}
\caption{Techniques used in the different modules of the LogAnswer system.}
\label{fig:modules}
\end{figure}
At first, information retrieval (IR), i.e. pattern matching, is used
to filter text passages suitable for the given question from the textual
representation of the Wikipedia. For this, the text sources are segmented into
sentence-sized passages. The corresponding representation can be enriched by the
descriptors of other sentences that result from a coreference solution of
pronouns, where the referred phrase is added to the description, if \eg\ the
pronoun 'he' refers to the individual 'Ian Fleming'. Then decision tree learning
ranks the text passages and chooses a set of answer candidates from these text
passages (Ranking step in Figure~\ref{fig:architecture}). Here, features like
the number of matching lexemes between passages and the question or the
occurrences of proper names in the passage are computed. Up to 200 text passages
are extracted from the knowledge base according to this ranking.

In the next step (Reasoning), the Hyper theorem prover is used to check if these
text passages provide an answer to the question. For every answer candidate, a
first-order logic representation of both the question and the answer candidate
is combined with a huge background knowledge. These proofs provide the answer to
the question by means of variable assignments. The proofs for the answer
candidates are then ranked again using decision tree learning (in the answer validation
phase). For the five best answers, text passages providing the answer are
highlighted and presented to the user. This is done by a natural language (NL)
answer generation module, which eventually yields the final answer candidates,
in our case that Ian Fleming is a British author.

In the LogAnswer system, various techniques work interlocked. See
Figure~\ref{fig:modules} for an overview of the different techniques together
with the modules in which they are used. Extraction of text passages for a
certain question is performed in the candidate selection module. In this module,
both information retrieval and decision tree learning work hand in hand to find
a list of answer candidates for the current question. For each answer candidate,
the reasoning module is invoked. This module consists of the Hyper theorem
prover, which is used to check if the answer candidate provides an answer for
the question. Since Hyper is able to handle first-order logic with equality and
knowledge bases given in description logic, it is possible to incorporate
background knowledge given in various (formal) languages.

An interesting extension of usual background knowledge is the use of a knowledge
base containing normative statements formalized in deontic logic. These
normative statements enable the system to reason in a rational way. Since
deontic logic can be translated into description logics, Hyper can be used to
reason on such knowledge bases. Reasoning in defeasible logic is another
technique contained in the reasoning module of the LogAnswer system. With the
help of defeasible logic reasoning, different proofs produced by Hyper are
compared. The proofs found by Hyper provide answers to the given question by
means of variable assignments. Comparing the proofs for different answer
candidates therefore is used to determine the best answer. Hence defeasible
logic is contained in the answer validation module. In addition to that,
the answer validation module contains decision tree learning to rank different
proofs found by Hyper and case-based reasoning. Details on the use of case-based
reasoning and reasoning in defeasible logic, that can both be used in the answer
validation phase (see Figure~\ref{fig:architecture}), can be found in
the Section~\ref{Searching for Good Answers}.

\section{Searching for Good Answers}\label{Searching for Good Answers}

As depicted before, the reasoning component of the LogAnswer system delivers
proofs, which represent the possible answers to the given question. The proofs
are ranked by decision trees which take into account several attributes of the
reasoning process together with the attribute from the previous information
retrieval step.

In addition to this ranking, we experiment with different other techniques to
improve the evaluation of answers. These are case-based reasoning (CBR)
(Section~\ref{similarity}), defeasible reasoning (Section~\ref{specificity}),
and normative (deontic) reasoning (Section~\ref{deontic}). To perform systematic
and extensive tests with LogAnswer, we used the CLEF database, strictly
speaking, its question answering part. CLEF stands for cross-language evaluation
forum, see \url{www.clef-campaign.org}. It is an international campaign
providing language data in different languages, \eg\ from newspaper articles.
Its workshop and competition series contains a track on question answering. We
used data from CLEF-2007 and CLEF-2008 \cite{giampiccolo2008overview,CLEF08}.

\subsection{CBR Similarity Measures and Machine Learning}
\label{similarity}

Answer validation can be enhanced by using experience knowledge in form of cases
in a case base. The resulting system module is designed as a learning system and
based on a dedicated CBR control structure. Contrary to common procedures in
natural-language processing, however, we do not follow the textual approach,
where experiences are available in unstructured or semi-structured text form,
but use a structured approach along the lines of \cite{Be02}. This is possible
because the knowledge source is available not only in textual but also in a
logical format. The semantics of the natural-language text is given basically by
first-order predicate logic formulae, which are represented by MultiNet graphs
\cite{He06}. Our basis is a manually achieved classification for each pair of
question (from the CLEF 2007 and 2008 data) and answer candidate (from the
LogAnswer system) whether the answer candidate is a good one for the question.
In order to compare and to define a similarity measure of the MultiNet graphs,
we have developed a new graph similarity measure \cite{GlWe12,We13} which
improves other existing measures, \eg\ \cite{Be02,BM94}.

\begin{figure}[!htb]
\includegraphics[width=\columnwidth]{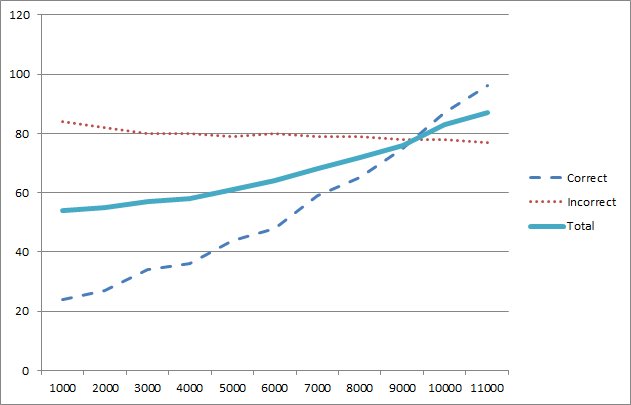}
\caption{The x-axis is the number of cases in the case base. The y-axis is the
classification accuracy in percent, for correct and incorrect answer candidates,
as well as the overall classification accuracy for the user interaction
simulation. \label{fig:userinteractionsimulation}}
\end{figure}

We measured the CBR system classification accuracy by running tests with a case
base from the CLEF 2007 and 2008 data. Our overall test set had 254 very 
heterogeneous questions and ca. 15000 cases. For instance, in one of 
the evaluations, namely the user interaction
simulation (see Figure~\ref{fig:userinteractionsimulation}), we examined the
development of the results for a growing knowledge base. We simulated users that
give reliable feedback to new, heterogeneous questions for which the LogAnswer system provides
answers candidates. The test setting was to guess the classification of
questions and answer candidates the system does not have in the knowledge base.
The results show the increase of the classification accuracy with a growing
number of correct cases in the case base. 
We performed a number of other evaluation experiments, \eg\ 3- and
10-fold cross validations. For more information about the integrated CBR/Machine
learning evaluation and test settings, please refer to \cite{GlWe12,We13}.

We further integrated case-based reasoning into the already existing answer
selection techniques in LogAnswer (answer validation phase, see Figure~\ref{fig:architecture}). For this, the results of the CBR stage were
turned into numeric features. A ranking model determined by a supervised
learning-to-rank approach combined these CBR-based features with other answer
selection features determined by shallow linguistic processing and logical
answer validation The final machine learning ranker is an ensemble of ten
rank-optimizing decision trees, obtained by stratified bagging, whose individual
probability estimates are combined by averaging. When training the machine
learning ranker on a case base optimized for perfect treatment of correct answer
candidates, we get the best overall result in our tests with a mean reciprocal
rank (MRR) of 0.74 (0.72 without CBR) and a correct top-ranked answer chosen in 61\% 
(58\% without CBR) of the cases.
It is instructive to consider the usage of CBR features in the machine learning
ranker, by inspecting all branching conditions in the generated trees and
counting the frequency of occurrence of each feature in such a branching
condition, since 10 bags of 10 decision trees were generated in the 10
cross-validation runs, there is a total of 100 trees to base results on
\cite{GlWe12,We13}. In total, 42.5\% of all split conditions in the learned
trees involve one of the CBR attributes. This further demonstrates the strong
impact of CBR results on answer re-ranking.

\subsection{The Specificity Criterion}
\label{specificity}

More specific answer candidates are to be preferred to less specific ones, and
we can compare them according to their specificity as follows. To obtain what
{\em argumentation theories}\/ call an {\em argument}, we form a pair of an
answer candidate and its derivation. The derivation can be based on {\em
positive-conditional rules}, generated from Hyper's verifications and capturing
the Wikipedia page of the answer candidate and the linguistic knowledge actually
applied. Now we find ourselves in the setting of {\em defeasible reasoning}\/
and can sort the arguments according to their specificity.
 
In defeasible reasoning, certain knowledge is assumed to be {\em defeasible}.
{\em Strict}\/ knowledge, however, is specified by {\em contingent facts} (\eg\
in the emu example from Section~\ref{introduction}, ``Tom is an emu'') and {\em
general rules}\/ holding in all possible worlds without exception (\eg\ ``emus
do not fly''). Strict knowledge is always preferred to knowledge depending also
on {\em defeasible rules} (\eg\ ``birds {\em normally}\/ fly'').

Already in 1985, \poolename\ had the idea to prefer more {\em specific}\/
arguments in case of conflicting results as follows
\cite{Poole-Preferring-Most-Specific-1985}: For any derivation of a given
result, represented as a tree, consider the sets of all leaves that contribute
to the applications of defeasible rules. An {\em activation set}\/ is a set of
literals from which all literals labeling such a set of leaves is derivable.
Thereby, an activation set is sufficient to activate the defeasible parts of a
derivation in the sense of a presupposition, {\em without using any additional
contingent facts}.

One argument is now {\em more specific}\/ than another one if all its activation
sets are activation sets of the other one. This means that each activation set
of the more specific argument (seen as the conjunction of its literals) must be
more specific than an activation set of the other one. Note that the meaning of
the latter usage of the word ``specific'' is just the {\em traditional common-sense
concept of specificity}, according to which a criterion (here: conjunction of
literals) is more specific than another one if it entails the other one.

We discovered several weaknesses of Poole's relation, such as its
non-transitivity: Contrary to what is obviously intended in
\cite{Poole-Preferring-Most-Specific-1985} and ``proved'' in
\cite{Simari-Loui-Defeasible-Reasoning-1992}, Poole's relation is not a
quasi-ordering and cannot generate an ordering. We were able to cure all the
discovered weaknesses by defining a {\em quasi-ordering}
\cite{Wirth_Stolzenburg_Specificity_KR2014,Wirth_Stolzenburg_Series_Specificity_AMAI_2015}
(\ie\ a reflexive and transitive binary relation), which can be seen as a
correction of Poole's relation, maintaining and clarifying Poole's original
intuition.

The intractability of Poole's relation, known at least since~2003
\cite{Stolzenburg-etal-Computing-Specificity-2003}, was attenuated by our
quasi-ordering and then overcome by restricting the rules to instances that were
actually used in the proofs found by Hyper, and by treating the remaining
variables (if any) as constants. With these restrictions, the intractability did
not show up anymore in any of the hundreds of examples we tested with our
\PROLOG\ \mbox{implementation}.

Running this implementation through the entire CLEF-2008 database, almost all
suggested answer solutions turned out to be incomparable \wrt\ specificity,
although our quasi-ordering can compare more arguments in practice than Poole's
original relation. One problem here is that we have to classify the rules of the
CLEF examples as being either general or defeasible, but there is no obvious way
to classify them. Another problem with the knowledge encoded in the MultiNet
formalism is that it first and foremost encodes only linguistic knowledge, \eg,
who is the agent of a given sentence. Only little background knowledge is
available, such as on ontology. All data from the web pages, however, are
represented by literals.

To employ more (defeasible) background knowledge we investigated other examples,
such as the emu example from Section~\ref{introduction}. Here, the formalization
in first-order logic of the natural-language knowledge on individuals can be
achieved with the Boxer system \cite{Bos2005IWCS,CurranClarkBos2007ACL}, which
is dedicated to large-scale language processing applications. These examples can
be successfully treated with the specificity criterion and also with deontic
logic (see subsequent section).

\subsection{Making Use of Deontic Logic}
\label{deontic}

Normative statements like ``you ought not steal'' are omnipresent in our
everyday life, and humans are used to do reason with respect to them. Since norms
can be helpful to model rationality, they constitute an important aspect for
common-sense reasoning. This is why normative reasoning is investigated in the
RatioLog project \cite{FSS14c}. Standard deontic logic (SDL)
\cite{gabbay2013handbook} is a logic which is very suitable for the
formalization of knowledge about norms. SDL corresponds to the modal logic
$\mathsf{K}$ together with a seriality axiom. In SDL the modal operator $\Box$
is interpreted as ``it is obligatory that'' and the $\Diamond$ operator as ``it
is permitted that''. For example a norm like ``you ought not steal'' can be
intuitively formalized as $\Box \lnot \mathit{steal}$. From a model theoretic
point of view, the seriality axiom contained in SDL ensures that, whenever it is
obligatory that something holds, there is always an ideal world fulfilling the
obligation.

In the RatioLog project, we experiment with SDL by adding normative statements
into the background knowledge. The emu example from Section~\ref{introduction}
contains the normative assertion
\begin{quote}\em
  Birds normally fly.
\end{quote}
which can be modeled using SDL as
\begin{equation*}
\mathit{Bird} \rightarrow \Box \mathit{Flies}
\end{equation*}
and is added to the background knowledge. In addition to normative statements,
the background knowledge furthermore contains assertions not containing any
modal operators, \eg\ something like the statement that all emus are birds.
Formulae representing contingent facts, like the assertion 
\begin{quote}\em
  Tom is an emu.
\end{quote}
in the emu example, are combined with the background knowledge containing
information about norms. The Hyper theorem prover \cite{cadesd} can be used to
analyze the resulting knowledge base. For example, it is possible to ask the
prover if the observed world with the emu Tom fulfills the norm that birds
usually are able to fly. 

Within the RatioLog project both defeasible logic and deontic logic are used.
There are similarities between defeasible logic and deontic logic. For example
in defeasible logic there are rules which are considered to be not strict but
defeasible. These defeasible rules are similar to normative statements, since
norms only describe how the world ought to be and not how it actually is. This
is why we are also investigating the connection between these two logics within
the RatioLog project.

\section{Conclusions}
\label{conclusion}

Deep question answering does not only require pattern matching and indexing
techniques, but also rational reasoning. This has been investigated within the
RatioLog project as demonstrated in this article. Techniques from machine
learning with similarity measures and case-based reasoning, defeasible reasoning
with (a revision of) the specificity criterion, and normative reasoning with
deontic logic help to select good answer candidates. If the background
knowledge, however, mainly encodes linguistic knowledge ---~without general
common-sense world knowledge~--- then the effect on finding good answer
candidates is low. Therefore, future work will concentrate on employing even
more background world knowledge (\eg\ from ontology databases), so that rational
reasoning can be exploited more effectively when applied to this concrete
knowledge.

\begin{acknowledgements}
The authors gratefully acknowledge the support of the DFG under the grants
FU~263/15-1 and STO~421/5-1 \emph{Ratiolog}.
\end{acknowledgements}

\bibliographystyle{spmpsci}
\bibliography{joint_paper.bib}

\pagebreak

\cv{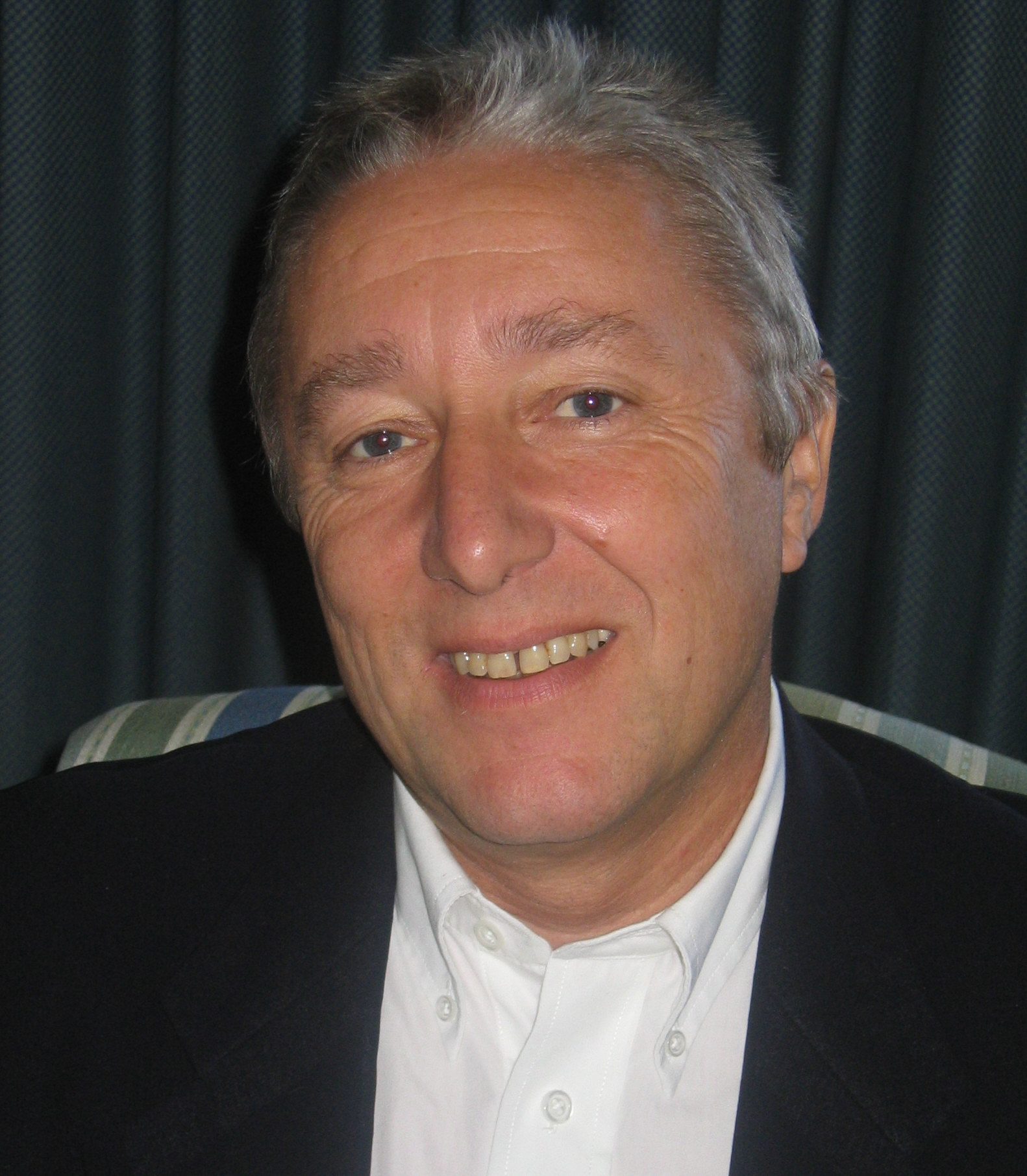}{Ulrich Furbach}{is a Senior Research Professor of Artificial
Intelligence at the University of Koblenz. His research interests include
knowledge management, automated reasoning, multiagent systems, and cognitive
science.}

\cv{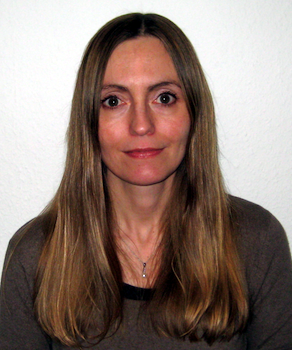}{Claudia Schon}{is employed at the University of Koblenz-Landau in the
RatioLog project. Her research interests include logic, especially description
logics, artificial intelligence, and cognition.}

\cv{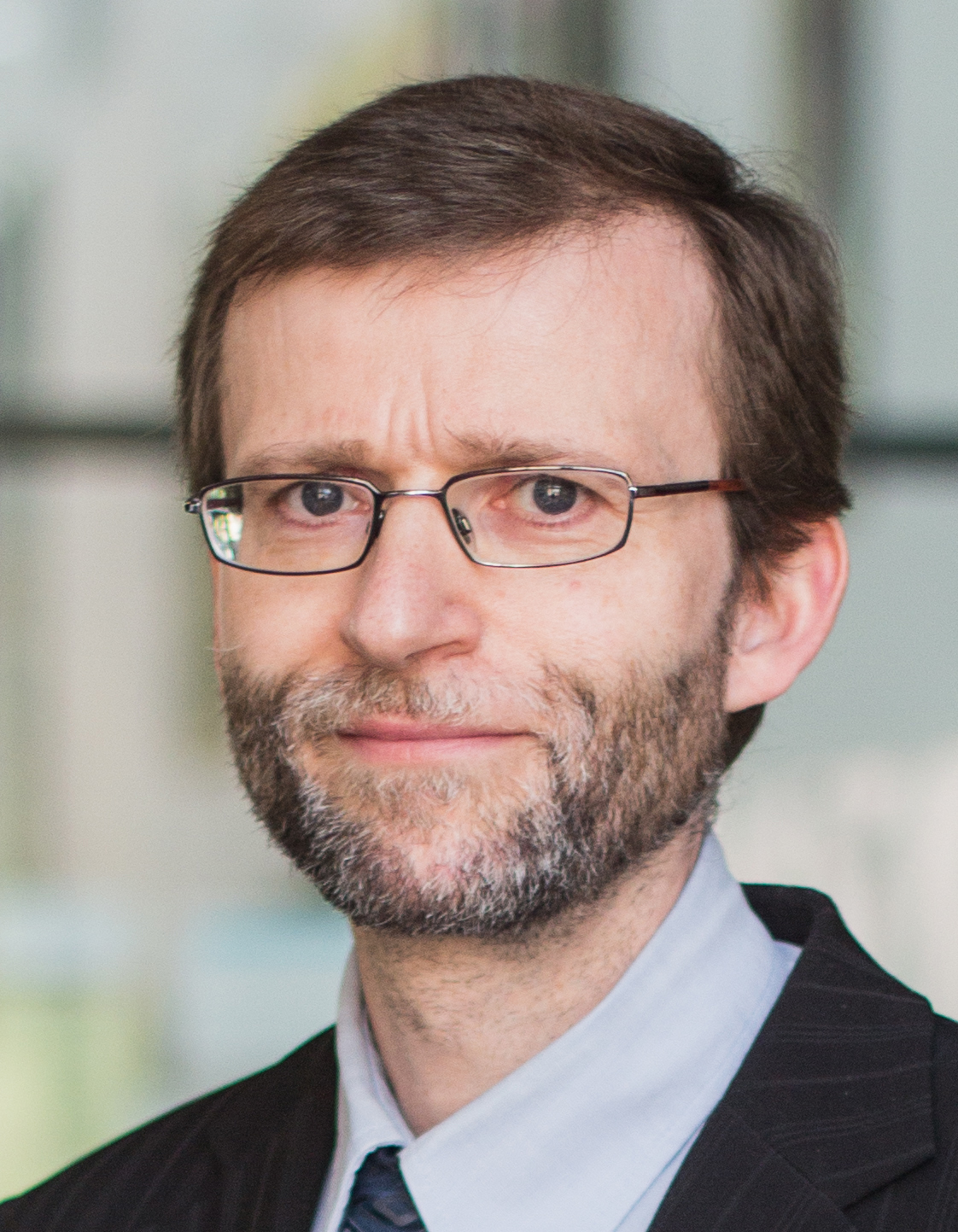}{Frieder Stolzenburg}{is professor of knowledge-based
systems at the Harz University of Applied Sciences in Wernigerode and directs
the Mobile Systems Laboratory at the Automation and Computer Science Department.
His research interests include artificial intelligence, knowledge representation
and logic, cognition and cognitive systems, and mobile robotics.}

\newpage

\cv{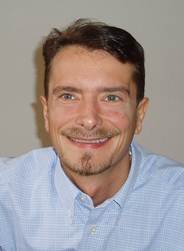}{Karl-Heinz Weis}{is director of his company Weis Consulting e.K., founded in 2006.
He permanently kept contact to the scientific
community of artificial intelligence while working for the industry.
His publications
include works about case-based reasoning, experience management, machine
learning, and neural networks.}

\cv{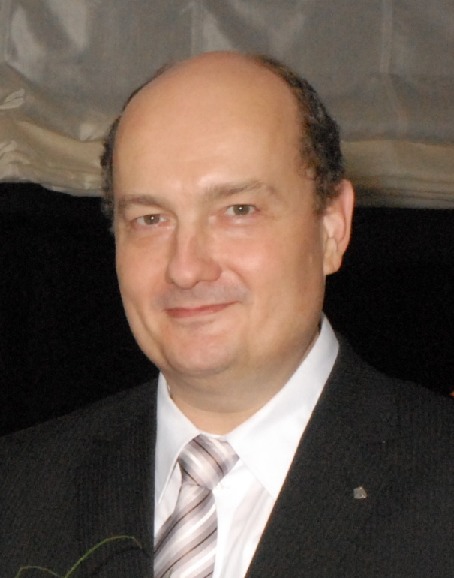}{Claus-Peter Wirth}{(Dr.\,rer.\,nat.)
has a half-time position in the RatioLog project 
and
is translator and editor in chief in the Hilbert--Bernays Project.
He was a member of
several SFBs, a SICSA distinguished visting fellow, and a 
visiting senior research 
fellow at King's College London. His research interests include 
artificial intelligence and the history of logic and philosophy.}

\bigskip\noindent The final publication is available at:\\
\url{http://link.springer.com/article/10.1007/s13218-015-0377-9}

\end{document}